\def\year{2019}\relax
\documentclass[letterpaper]{article} 
\usepackage{xcolor}
\usepackage[hyphens]{url}  \Urlmuskip=0mu plus 1mu
\RequirePackage[colorlinks=true,allcolors=blue!50!gray]{hyperref}
\usepackage{aaai19}  
\usepackage{times, helvet, courier}
\RequirePackage{adjustbox}
\RequirePackage{amsmath,amsfonts,amssymb}
\RequirePackage{enumitem} \setlist{nosep}
\setlist[description]{font={\bfseries\slshape\color{gray!90!red}}}
\usepackage[textsize=scriptsize,backgroundcolor=red!10,bordercolor=red!90]{todonotes}
\makeatletter
\g@addto@macro{\normalsize}{%
    \setlength{\abovedisplayskip}{0pt}
    \setlength{\abovedisplayshortskip}{0pt}
    \setlength{\belowdisplayskip}{0pt}
    \setlength{\belowdisplayshortskip}{0pt}
    \setlength{\textfloatsep}{.9ex plus.5ex}}
\makeatother
\RequirePackage[compact]{titlesec}
\titlespacing{\subsubsection}{0pt}{.6ex plus .1ex minus .2ex}{1em}
\titlespacing{\paragraph}{0pt}{.4ex plus .1ex minus .3ex}{1em}
\RequirePackage[normalem]{ulem}
\RequirePackage{soul}
\RequirePackage{colortbl}
\soulregister{\underline}{1}
\def\yhl#1#2{{\definecolor{tempcolor}{rgb}{#1}\sethlcolor{tempcolor}\hl{#2}}}
\RequirePackage{array}
\RequirePackage{multirow}
\RequirePackage[skip=.3ex plus.1ex]{caption}
\RequirePackage{flushend}
\RequirePackage{natbib}
\RequirePackage[most]{tcolorbox}
\RequirePackage{tikz} \usetikzlibrary{arrows}
\usepackage{pifont}

\def\longname{Gated Interleaved Recurrent Network}
\def\shortname{GIRNet}
\def\system{\shortname}

\begin{document}
\title{GIRNet: Interleaved Multi-Task Recurrent State Sequence Models}

\author{Divam Gupta,\textsuperscript{\ding{81}} Tanmoy Chakraborty,\textsuperscript{\ding{81}} Soumen Chakrabarti\textsuperscript{\ding{169}}\\
\textsuperscript{\ding{81}}IIIT Delhi, India, \textsuperscript{\ding{169}}IIT Bombay, India\\
\{divam14038, tanmoy\}@iiitd.ac.in, soumen@cse.iitb.ac.in}

\maketitle

\begin{abstract}
In several natural language tasks, labeled sequences are available in separate domains (say, languages), but the goal is to label sequences with mixed domain (such as code-switched text).  Or, we may have available models for labeling whole passages (say, with sentiments), which we would like to exploit toward better position-specific label inference (say, target-dependent sentiment annotation).  A key characteristic shared across such tasks is that different positions in a primary instance can benefit from different `experts' trained from auxiliary data, but labeled primary instances are scarce, and labeling the best expert for each position entails unacceptable cognitive burden.  We propose \shortname, 
a unified position-sensitive multi-task recurrent neural network (RNN) architecture for such applications.  Auxiliary and primary tasks need not share training instances.  Auxiliary RNNs are trained over auxiliary instances.  A primary instance is also submitted to each auxiliary RNN, but their state sequences are gated and merged into a novel composite state sequence tailored to the primary inference task.  Our approach is in sharp contrast to recent multi-task networks like the cross-stitch and sluice network, which do not control state transfer at such fine granularity.  We demonstrate the superiority of \shortname\ using three applications: sentiment classification of code-switched passages, part-of-speech tagging of code-switched text, and target position-sensitive annotation of sentiment in monolingual passages.  In all cases, we establish new state-of-the-art performance beyond recent competitive baselines.


\end{abstract}


\section{Introduction}
\label{sec:Intro}

Neural networks have shown outstanding results in many Natural Language Processing (NLP) tasks, particularly involving sequence labeling \citep{schmid1994npos,lample2016NeuralNER} and sequence-to-sequence translation \citep{BahdanauCB2014Attention}.  The success is generally attributed to their ability to learn good representations and recurrent models in an end-to-end manner.  Most deep models require generous volumes of training data to adequately train their large number of parameters.  Collecting sufficient labeled data for some tasks entails unacceptably high cognitive burden.  To overcome this bottleneck, some form of transfer learning, semi-supervised learning or multi-task learning (MTL) is used.

In the most common form of MTL, related tasks such as part-of-speech (POS) tagging and named entity recognition (NER) share representation close to the input, e.g., as character n-gram or word embeddings, followed by separate networks tailored to each prediction task \citep{Sogaard2016MTLNLP,MaurerPR2016MTL}.  Care is needed to jointly train shared model parameters to prevent one task from hijacking the representation \citep{Ruder2017MTLsurvey}.

Sequence labeling brings a new dimension to MTL.  In many NLP tasks, labeled sequences are readily available in separate single languages, but our goal may be to label sequences from code-switched multilingual text.  Or, we may have trained models for labeling whole sentences or passages with overall sentiment, but the task at hand is to infer the sentiment expressed toward a specific entity mentioned in the text.  Ideally, we would like to build a composite state sequence representation where each position draws state information from the best `expert' auxiliary sequence, but instances are diverse in terms of where the experts switch, and anyway annotating these transitions would entail prohibitive cognitive cost.    We survey related work in Section~\ref{sec:Rel}, and explain why most of them do not satisfy our requirements.

By abstracting the above requirements into a unified framework, in Section~\ref{sec:Arch} we present \shortname{} ({\bf G}ated {\bf I}nterleaved {\bf R}ecurrent {\bf Net}work)\footnote{\raggedright The code is available at \url{https://github.com/divamgupta/mtl\_girnet} }, a novel MTL network tailored for dynamically daisy-chaining at a \emph{word level} the best experts along the input sequence to derive a composite state sequence that is ideally suited for the primary task.  The whole network over all tasks is jointly trained in an end-to-end manner.  \shortname{} applies both when we have common instances for the tasks and when we have only disjoint instances over the tasks.  Motivated by low-resource NLP challenges, we assume that one task is \emph{primary} but with scant labeled data, whereas the \emph{auxiliary} tasks have more labeled data available.  We train standard LSTMs on the auxiliary models; but when a primary instance is run through the auxiliary networks, their states are interleaved using a novel, dynamic gating mechanism.  Remarkably, without access to any segmentation of primary instances, \shortname{} learns gate values such that auxiliary states are daisy-chained to maximize primary task performance.

We can think of \shortname{} as emulating an LSTM whose cell unit changes dynamically with input tokens.  E.g., for each token in a code-switched sentence over two languages, \shortname{} learns to choose the cell unit from an LSTM trained on text in the language in which that token is written.  \shortname{} achieves this with an end-to-end differentiable network that does not need supervision about which language is used for each token.

In Section~\ref{sec:Concrete}, we instantiate the \shortname{} template to three concrete applications: sentiment labeling of code-switched passages, POS tagging of code-switched passages, and target-dependent (monolingual) sentiment classification.  Where the primary task is to tag code-switched multilingual sequences, the auxiliary tasks would be to tag the component monolingual texts.  For target-dependent sentiment classification, the auxiliary task would be target-independent whole-passage sentiment classification.  In all three applications, we consistently beat competitive baselines to establish new state-of-the-art performance.  These experiments are described in Section~\ref{sec:Expt}.

Summarizing, our contributions are four-fold:
\begin{itemize}
\item \shortname, a novel position- and context-sensitive dynamic gated recurrent network to compose state sequences from auxiliary recurrent units.
\item Three applications that are readily expressed as concrete instantiations of the \shortname{} framework.
\item Superior performance in all three applications.
\item Thorough diagnostic interpretation of \shortname's behavior, demonstrating the intended gating effect.
\end{itemize}

\section{Related work}
\label{sec:Rel}

The most common neural MTL architecture shares parameters in initial layers (near the inputs) and train separate task-specific layers for per-task prediction \citep{Caruana93,Sogaard2016MTLNLP,MaurerPR2016MTL}. 
In applications where tasks are not closely related, finding a common useful representation for all tasks is hard.  Moreover, jointly training shared model parameters, while preventing one task from hijacking the representation, may be challenging \citep{Ruder2017MTLsurvey}. In soft parameter sharing \citep{duong2015low}, each task has its separate set of parameters and the distance between the inter-task parameters is minimized by adding an additional loss while training. 

In some cases, rather than just sharing the parameters (completely or partially),(state sequence) features extracted by the model for one task are fed into the model of another task.
In shared-private MTL models, there is one common shared model over all tasks and also separate private models for each task.
The following approaches combine information from shared and private LSTMs at the granularity of token positions.  \citet{liu2017adversarial} and \citet{Chen2018MetaMTL} use one shared LSTM and one private LSTM per task.  \citet{liu2017adversarial} concatenate their outputs, whereas \citet{Chen2018MetaMTL} concatenate the shared LSTM state with the input embeddings. 
In the low-supervision MTL model \citep{Sogaard2016MTLNLP}, auxiliary tasks are trained on the lower layers and the primary task is trained on the higher layer. 
In all these models, there is no control over amount of information shared between different tasks.
To overcome this problem, various network architectures have been evolved to more carefully control the transfer across different tasks.
Cross-stitch \citep{Misra2016CrossStitch} and sluice \citep{Ruder2017sluice} networks are two such frameworks. 
\citet{Chen2018RLMTL} have used reinforcement learning to search for the best patterns of sharing between tasks.  However, transfer happens at the granularity of layers, and not recurrent positions.  Also, transfer happens usually via vector concatenation.  No gating information crosses RNN boundaries.  
Meta-MTL \citep{Chen2018MetaMTL} uses an LSTM shared across all tasks to control the parameters of the task-specific LSTMs.
As will become clear, we gain more representational power by gating and interleaving of auxiliary states driven by the input.  Further details of some of the above approaches are discussed along with experiments in Section~\ref{sec:Expt}.

\section{Formulation and proposed architecture}
\label{sec:Arch}

Our abstract problem setting consists of a \textbf{primary} task and $m\ge 1$ \textbf{auxiliary} tasks.  E.g., the primary task may be part-of-speech (POS) tagging of code-mixed (say, English and Spanish) text, and the two auxiliary tasks may be POS tagging of pure English and Spanish text.  All tasks involve labeling sequences, but the labels may be per-token (e.g., POS tagging) or global (e.g., sentiment analysis).  Labeled data for the primary task is generally scarce compared to the auxiliary tasks. Different tasks will generally have disjoint sets of instances.  Our goal is to mitigate the paucity of primary labeled data by transferring knowledge from auxiliary labeled data.  Our proposed architecture is particularly suitable when different parts or spans of a primary instance are related to different auxiliary tasks, and these primary spans are too expensive or impossible to identify during training.

In this section, we introduce \shortname{} (\longname), our deep multi-task architecture which learns to skip or select spans in auxiliary RNNs based on its ability to assist the primary labeling task.  We describe the system by using LSTMs, but similar systems can be built using other RNNs such as GRUs.  Before going into details, we give a broad outline of our strategy.  Auxiliary labeled instances are input to auxiliary LSTMs to reduce auxiliary losses (Section~\ref{sec:AuxAux}).  Each primary instance is input to a `gating' LSTM (Section~\ref{sec:Gate}) and a variation of each auxiliary LSTM, which run concurrently with a composite sequence assembled from the auxiliary state sequences using the scalar gate values (Section~\ref{sec:PrimAuxComp}).  In effect, our network learns to dynamically daisy-chain the best auxiliary experts word by word.  Finally, the composite state sequence is combined with a primary LSTM state sequence to produce the primary prediction, with its corresponding primary loss (Section~\ref{sec:PrimPred}).

\begin{figure}[th]
  \tikzset{ >=latex, auto }
  \tikzstyle{rnncell}=[rectangle, draw, fill=gray!15, align=center,
    minimum height=10mm, minimum width=14mm]
  \tikzstyle{jnct}=[circle, fill=black, minimum size=1mm, inner sep=0]
  \tikzstyle{opjn}=[inner sep=0, outer sep=0]
  \centering
  \begin{tikzpicture}
    \node (dp0) {$c_{t-1}^{\text{comp}}$} ;
    \node (j0) [jnct, right=4mm of dp0] {} ;
    \node (dp1) [below=2mm of dp0] {$h_{t-1}^{\text{comp}}$} ;
    \node (j1) [jnct, right=6mm of dp1] {} ;
    \node (PrimRnn) [rnncell, below right=6mm and 13mm of dp1] {prim} ;
    \node (Aux1Rnn) [rnncell, below=24mm of PrimRnn, draw] {$\text{aux}_1$} ;
    \node (Aux2Rnn) [rnncell, below=6mm of Aux1Rnn, draw] {$\text{aux}_2$} ;
    \draw [->] (j1) |- node[below, pos=.9] {$h$} (Aux1Rnn.195) ;
    \draw [->] (j1) |- node[below, pos=.9] {$h$} (Aux2Rnn.195) ;
    \draw [->] (j0) |- node[above, pos=.9] {$c$} (Aux1Rnn.165) ;
    \draw [->] (j0) |- node[above, pos=.9] {$c$} (Aux2Rnn.165) ;
    \node (ct1prim) [left=12mm of PrimRnn.150] {$c_{t-1}^{\text{prim}}$} ;
    \path [->] (ct1prim) edge (PrimRnn.150) ;
    \node (ht1prim) [left=12mm of PrimRnn.210] {$h_{t-1}^{\text{prim}}$} ;
    \path [->] (ht1prim) edge (PrimRnn.210) ;
    \node (ctprim) [right=35mm of PrimRnn.30] {$c_{t}^{\text{prim}}$} ;
    \path [->] (PrimRnn.30) edge (ctprim) ;
    \node (htprim) [right=35mm of PrimRnn.330] {$h_{t}^{\text{prim}}$} ;
    \path [->] (PrimRnn.330) edge (htprim) ;
    \node (j2) [jnct, left=3mm of PrimRnn.210] {} ;
    \node (phi) [below right=10mm and 3mm of PrimRnn,
    minimum width=13mm, draw, fill=yellow!10] {$\phi$} ;
    \draw [->] (j2) |- (phi) ;
    \node (h1odot) [opjn] at (phi.230 |- Aux1Rnn.15) {$\odot$} ;
    \node (h2odot) [opjn] at (phi.230 |- Aux2Rnn.15) {$\odot$} ;
    \node (c1odot) [opjn] at (phi.310 |- Aux1Rnn.345) {$\odot$} ;
    \node (c2odot) [opjn] at (phi.310 |- Aux2Rnn.345) {$\odot$} ;
    \path [->] (Aux1Rnn.15) edge (h1odot) ;
    \path [->] (Aux2Rnn.15) edge (h2odot) ;
    \path [->] (Aux1Rnn.345) edge (c1odot) ;
    \path [->] (Aux2Rnn.345) edge (c2odot) ;
    \path [->] (phi.230) edge (h1odot) ;
    \path [->] (phi.230) edge [bend right]
    node [left, pos=0.1] {$g_t[1]$} (h2odot) ;
    \path [->] (phi.310) edge (c1odot) ;
    \path [->] (phi.310) edge [bend left]
    node [right, pos=0.1] {$g_t[2]$} (c2odot) ;
    \node (compodot) [opjn] at (dp1 -| phi){$\odot$} ;
    \node (compoplus) [opjn, right=12mm of compodot] {$\bigoplus$} ;
    \node (ccompodot) [opjn] at (dp0 -| phi) {$\odot$} ;
    \node (ccompoplus) [opjn, right=16mm of ccompodot] {$\bigoplus$} ;
    \path [->] (phi.north) edge
    node[left,pos=0.1] {$1-\sum_j g_t[j]$} (compodot) ;
    \path [->] (phi.north) edge [bend right] (ccompodot) ;
    \node (dq0) [right=6mm of ccompoplus] {$c_{t}^{\text{comp}}$} ;
    \path [->] (dp0) edge (ccompodot) ;
    \path [->] (ccompodot) edge (ccompoplus) ;
    \path [->] (ccompoplus) edge (dq0) ;
    \node (dq1) [right=10mm of compoplus] {$h_{t}^{\text{comp}}$} ;
    \path [->] (dp1) edge (compodot) ;
    \path [->] (compodot) edge (compoplus) ;
    \path [->] (compoplus) edge (dq1) ;
    \draw [->] (h1odot) -| (ccompoplus.250) ;
    \draw [->] (h2odot) -| (ccompoplus.290) ;
    \draw [->] (c1odot) -| (compoplus.250) ;
    \draw [->] (c2odot) -| (compoplus.290) ;
  \end{tikzpicture}
\caption{Part of \system\ that processes primary input $x_t^{\text{prim}}$, which is provided to primary LSTM `prim' and auxiliary LSTMs `$\text{aux}_j$' for $j=1,2$ in this example, as well as gating logic $\phi$. ($x_t^{\text{prim}}$ has been elided to reduce clutter.)  Training of auxiliary LSTMs on auxiliary inputs is standard, and has been omitted for clarity.  $h_t^{\text{comp}}$ is the gated composite state sequence.  $\oplus$ represents elementwise addition and $\odot$ represents elementwise multiplication of the vector input with the scalar gate input.}
\label{fig:MTTDSCnetwork0}
\end{figure}
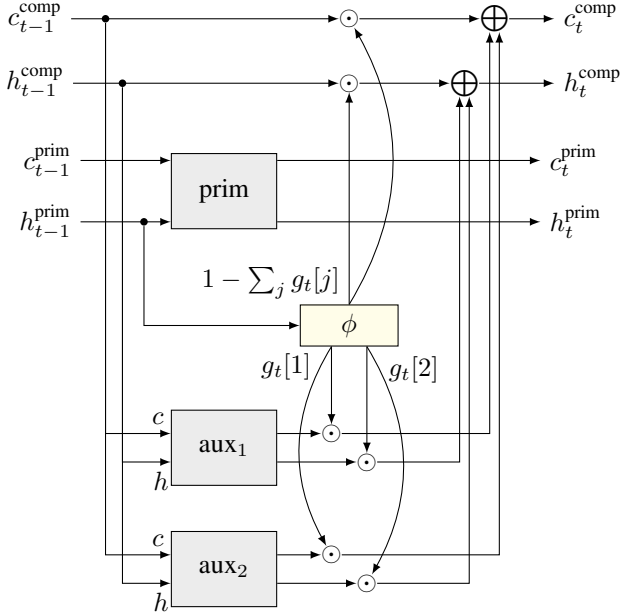

Auxiliary and primary losses are jointly optimized.  Weights of each auxiliary LSTM are updated by its own loss function and the loss of the primary task, but not any other auxiliary task.  Remarkably, the scalar gating LSTM can learn \emph{without supervision} how to assemble the composite state sequence.  Qualitative inspection of the gate outputs show that the selection of the auxiliary states indeed gets tailored to the primary task.  A high-level sketch of our architecture is shown in Figure~\ref{fig:MTTDSCnetwork0} with two auxiliary tasks and one primary task.

\subsection{Input embedding layer}
\label{sec:Input}

Each input instance is a sentence- or passage-like sequence of words or tokens that are mapped to integer IDs.  An embedding layer maps each ID to a $d$-dimensional vector.  The resulting sequence of embedding vectors for a sentence will be called~$x$.  We use a common embedding matrix over all auxiliary and primary tasks.  The embedding matrix is initialized randomly or using pre-trained embeddings \citep{PenningtonSM2014GloVe,MikolovSCCD2013word2vec}, and then trained along with the rest of our network.  Labeled instances will be accompanied by a suitable label~$y$.

\subsection{Auxiliary LSTMs run on auxiliary input}
\label{sec:AuxAux}

For each auxiliary task $j \in [1,m]$, we train a separate auxiliary LSTM using a separate set of labeled instances, which we call $x^{\text{auxaux}_j} = (x_1^{{auxaux}_j}, \ldots, x_n^{{auxaux}_j})$, and the ground-truth as  $y^{\text{auxaux}_j}$.  The signature and dimension of $y^{\text{auxaux}_j}$ can vary with the nature of the auxiliary task (classification, sequence labeling, etc.) 

For auxiliary task $j$, we define the LSTM state at each time step $t$: input gate $i_t^{\text{auxaux}_j}$, output gate $o_t^{\text{auxaux}_j}$, forget gate $f_t^{\text{auxaux}_j}$, memory state $c_t^{\text{auxaux}_j}$ and hidden state $h_t^{\text{auxaux}_j}$, all vectors in $ \mathbb{R}^d$.  For all auxiliary task models, we use the same number of hidden units.  The weights written in blue indicate that they are also used for the primary input, described later.

\begin{align}
\begin{bmatrix}
\widetilde{c}_t^{\text{auxaux}_j} \\
o_t^{\text{auxaux}_j} \\
i_t^{\text{auxaux}_j} \\
f_t^{\text{auxaux}_j} 
\end{bmatrix}
&= 
\begin{bmatrix}
\tanh{} \\
\sigma \\
\sigma \\
\sigma 
\end{bmatrix}
\left( \textcolor{blue}{W^{\text{aux}_j}} \begin{bmatrix} x_t^{\text{auxaux}_j} \\ h_{t-1}^{\text{auxaux}_j} \end{bmatrix} \right) \\
c_t^{\text{auxaux}_j} &= \widetilde{c}_t^{\text{auxaux}_j} \odot i_t^{\text{auxaux}_j} + c_{t-1}^{\text{auxaux}_j} \odot c_t^{\text{auxaux}_j}\\
h_t^{\text{auxaux}_j} &= o_t^{\text{auxaux}_j} \odot \tanh{c_t^{\text{auxaux}_j}}
\end{align}

(For compact notation we have written the operators to be applied in a column vector.) Using the hidden states of the LSTM for the auxiliary task, we get the desired output using another model $M^{\text{auxaux}_j}$.  E.g., we can use a fully connected layer over the last hidden states or pooled hidden states for the whole-sequence classification. We can use a fully connected layer on each hidden state for sequence labeling.  Using generic notation,
\begin{align}
\text{out}^{\text{auxaux}_j} = M^{\text{auxaux}_j}( h_1^{\text{auxaux}_j} , h_2^{\text{auxaux}_j}, \dots, h_n^{\text{auxaux}_j} )
\end{align}
For each auxiliary task $j$, a separate loss 
\begin{align}
\text{loss}^{\text{auxaux}_j}( \text{out}^{\text{auxaux}_j} , y^{\text{auxaux}_j} )
\end{align}
is computed using the model output and the ground-truth.  The losses of all auxiliary tasks are added to the final loss.

\subsection{LSTM on primary instance to produce gating signal}
\label{sec:Gate}

A primary input instance is written as $x^{\text{prim}} = (x_1^{prim}, \ldots, x_n^{prim})$ and the ground-truth label as $y^{\text{prim}}$.  At each token position $t$, the gating LSTM has internal representations as follows: input gate $i_t^{\text{prim}}$, output gate $o_t^{\text{prim}}$, forget gate $f_t^{\text{prim}}$, memory state $c_t^{\text{prim}}$ and hidden state $h_t^{\text{prim}}$.  All these vectors are in $\mathbb{R}^{d\prime}$, where $d\prime$ is the number of hidden units in the gating LSTM.

\begin{align}
\begin{bmatrix}
\widetilde{c}_t^{\text{prim}} \\
o_t^{\text{prim}} \\
i_t^{\text{prim}} \\
f_t^{\text{prim}} 
\end{bmatrix}
&= 
\begin{bmatrix}
\tanh{} \\
\sigma \\
\sigma \\
\sigma 
\end{bmatrix}
\left( W^{\text{prim}} \begin{bmatrix} x_t^{\text{prim}} \\ h_{t-1}^{\text{prim}} \end{bmatrix} \right) \\
c_t^{\text{prim}} &= \widetilde{c}_t^{\text{prim}} \odot i_t^{\text{prim}} + c_{t-1}^{\text{prim}} \odot f_t^{\text{prim}} \\
h_t^{\text{prim}} &= o_t^{\text{prim}} \odot \tanh{c_t^{\text{prim}}}
\end{align}

Here we describe the RNN that produces the gating signal as a uni-directional LSTM, but, depending on the application, we could use bi-directional LSTMs.  Using the hidden state of the gating LSTM and its input at token position $t$, we compute gate vector $g_t \in \mathbb{R}^m$, where $m$ is the number of auxiliary tasks. I.e., for each auxiliary task, we predict a scalar gate value.

\begin{align}
g_t = \phi\left( W^{\text{gator}} \begin{bmatrix} x_t^{\text{prim}} \\ h_{t-1}^{\text{prim}} \end{bmatrix} \right)
\end{align}

Here $\phi$ is an activation function which should ensure $\sum_{j=1}^m g_t[j] \leq 1$.  The rationale for this stipulation will become clear shortly from Equations \eqref{eq:htcomp} and \eqref{eq:ctcomp}.  We implement $\phi$ by including a fully-connected layer that generates $m+1$ scalar values, followed by applying soft-max, and discarding the last value.

\subsection{Auxiliary and gated composite LSTMs run on primary input}
\label{sec:PrimAuxComp}

For prediction on a primary instance to benefit from auxiliary models, the primary instance is also processed by a variant of each auxiliary LSTM.  The model weights $W^{\text{aux}_j}$ of each auxiliary LSTM will be borrowed from the corresponding auxiliary task, but the input is $x_t^{\text{prim}}$ and the states will be composite states over all auxiliary LSTMs.  Therefore the internal cell variables will have different values, which we therefore give different names: input gate $i_t^{\text{primaux}_j}$, output gate $o_t^{\text{primaux}_j}$, forget gate $f_t^{\text{primaux}_j}$, memory state $c_t^{\text{primaux}_j}$ and hidden state $h_t^{\text{primaux}_j}$.  All these vectors are in $\mathbb{R}^d$.

A key innovation in our architecture is that, along with these auxiliary states, we will compute, position-by-position, a \emph{gated, composite} state sequence comprised of $c_t^{\text{comp}}$ and $h_t^{\text{comp}}$.  The idea is to draw upon that auxiliary task, if any, that is best qualified to lend representation to the primary task, position by position.  Recall from Section~\ref{sec:Gate} that $g_t[j]$ denotes the relevance of the auxiliary model $j$ at token position $t$ of the primary instance.  If $g_t[j]\to1$, then the state of auxiliary model $j$ strongly influences $h^{\text{comp}}$ in the next step.  If, for all $j$, $g_t[j]\to0$, then no auxiliary model is helpful at position $t$.  Therefore, the previous composite state is passed to the next time step as is.  The skip possibility also makes training easier by countering vanishing gradients.

\begin{align}
\textcolor{blue}{h_t^{\text{comp}}} &= \sum_{j=1}^{m} h_t^{\text{primaux}_j} g_t[j]  + \left( 1 - \sum_{j=1}^m g_t[j] \right) h_{t-1}^{\text{comp}} \label{eq:htcomp} \\
\textcolor{blue}{c_t^{\text{comp}}} &= \sum_{j=1}^{m} c_t^{\text{primaux}_j} g_t[j]  + \left( 1 - \sum_{j=1}^m g_t[j] \right) c_{t-1}^{\text{comp}} \label{eq:ctcomp}
\end{align}

E.g., in target-dependent sentiment analysis, the auxiliary LSTM may identify sentiment-bearing words irrespective of target, and the composite state sequence prepares the stage for detecting the polarity of sentiment at a designated target.

\begin{align}
\begin{bmatrix}
\widetilde{c}_t^{\text{primaux}_j} \\
o_t^{\text{primaux}_j} \\
i_t^{\text{primaux}_j} \\
f_t^{\text{primaux}_j} 
\end{bmatrix}
&= 
\begin{bmatrix}
\tanh{} \\
\sigma \\
\sigma \\
\sigma 
\end{bmatrix}
\left( \textcolor{blue}{W^{\text{aux}_j}} \begin{bmatrix} x_t^{\text{prim}} \\
\textcolor{blue}{h_{t-1}^{\text{comp}}} \end{bmatrix} \right) \\
c_t^{\text{primaux}_j} &= \widetilde{c}_t^{\text{primaux}_j} \odot i_t^{\text{primaux}_j} + \textcolor{blue}{c_{t-1}^{\text{comp}}} \odot f_t^{\text{primaux}_j} \\
h_t^{\text{primaux}_j} &= o_t^{\text{primaux}_j} \odot \tanh{c_t^{\text{primaux}_j}}
\end{align}

\begin{figure}[t]
\tikzset{ >=latex, auto, node distance=13mm, }
\centering
\begin{tikzpicture}
\node [fill=gray!30] (xaux) {$x^{\text{aux}}$};
\node (xauxhauxaux) [below of=xaux] {};
\node (hauxaux) [below of=xauxhauxaux] {$h^{\text{auxaux}}$};
\node (outauxaux) [left=4mm of hauxaux] {$\text{out}^{\text{auxaux}}$};
\path [->] (hauxaux) edge (outauxaux);
\path (hauxaux) edge [loop below] (hauxaux);
\path [->] (xaux) edge (hauxaux);
\node [fill=gray!30] (xprim) [right of=xaux] {$x^{\text{prim}}$};
\node (xprimhprimaux) [below of=xprim] {};
\path (xauxhauxaux) edge [dashed] node[above]
{\textcolor{blue}{$W^{\text{aux}}$}} (xprimhprimaux);
\node (hprimaux) [right of=hauxaux] {$h^{\text{primaux}}$};
\path (hprimaux) edge [loop below] (hprimaux);
\path [->] (xprim) edge (hprimaux);
\node (hprim) [right of=xprim] {$h^{\text{prim}}$};
\path [->] (xprim) edge (hprim);
\path (hprim) edge [loop above] (hprim);
\node (mprim) [right of=hprim] {$M^{\text{prim}}$};
\node (outprim) [right=4mm of mprim] {$\text{out}^{\text{prim}}$};
\node (dummy2) [below of=mprim] {};
\node (gate) [below of=hprim] {$g$};
\path [->] (hprim) edge (gate);
\path [->] (xprim) edge (gate);
\node (hAUX) [below of=dummy2] {\textcolor{blue}{$h^{\text{comp}}$}};
\path (hAUX) edge [loop below] (hAUX);
\path [->] (hprimaux.10) edge [bend left] (hAUX.170);
\path [->] (hAUX.190) edge [bend left] (hprimaux.350);
\path [->] (gate) edge (hAUX);
\path [->] (hAUX) edge (mprim);
\path [->] (hprim) edge (mprim);
\path [->] (mprim) edge (outprim);
\node (lossauxaux) [above of=outauxaux] {$\text{loss}^{\text{auxaux}}$};
\node [fill=gray!30] (yaux) [above of=lossauxaux] {$y^{\text{aux}}$};
\path [->] (outauxaux) edge (lossauxaux);
\path [->] (yaux) edge (lossauxaux);
\node (lossprim) [below of=outprim] {$\text{loss}^{\text{prim}}$};
\node [fill=gray!30] (yprim) [below of=lossprim] {$y^{\text{prim}}$};
\path [->] (outprim) edge (lossprim);
\path [->] (yprim) edge (lossprim);
\end{tikzpicture}
\caption{Variables defined using other variables and observed constants.  Auxiliary and primary tasks are tied via~$W^{\text{aux}}$.  Self-loops indicate recurrence.  Only states $h$ shown for simplicity; $c$ assumed to accompany them.  } 
\label{fig:VarFlow}
\end{figure}
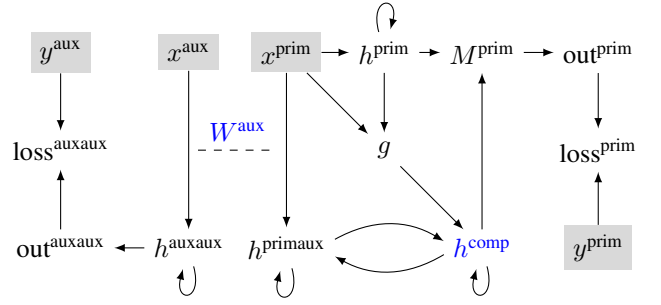

The overall flow of information between  many embedding variables is complicated. Therefore, we have sketched it for clarity in Figure~\ref{fig:VarFlow}.
To elaborate further, at step $t-1$, we compute $h_{t-1}^{\text{comp}}$ by combining all $h^{\text{primaux}}$ using gate signals.  The composite state is then fed into each  auxiliary cell in the current step~$t$.

\subsection{Combining gating and composite states for primary prediction}
\label{sec:PrimPred}

Using the hidden states from the axillary LSTMs and the primary LSTM, we get the desired output using another model $M^{\text{prim}}$. For example, we can use a fully connected layer over the last hidden states or pooled hidden states for classification. We can use a fully connected layer on all hidden states for sequence labeling tasks.

\begin{align}
\text{out}^{\text{prim}} = M^{\text{prim}}(h_1^{\text{prim}}, \ldots, h_n^{\text{prim}}; h_1^{\text{comp}}, \ldots, h_n^{\text{comp}})
\end{align}

The loss of the primary task,

\begin{align}
\text{loss}^{\text{prim}}(\text{out}^{\text{prim}} , y^{\text{prim}}),
\end{align}

is computed using the model output and the ground truth of the primary task data point.

\subsection{Training and regularization}

The system is trained jointly across auxiliary and primary tasks.  We sample $\{x^{\text{prim}}, y^{\text{prim}}\}$ from the primary task dataset and $\{x^{\text{auxaux}_1},y^{\text{auxaux}_1} \}, \{x^{\text{auxaux}_2},y^{\text{auxaux}_2} \}, \dots , \{x^{\text{auxaux}_m},y^{\text{auxaux}_m} \}$ from the auxiliary task datasets respectively. We compute the loss for the primary task and the auxiliary tasks as defined above. The total loss is the weighted sum of the loss of the primary task and the auxiliary tasks, where $\alpha_j$ is the weight of the auxiliary task.
\begin{align}
\text{loss}^{\text{all}} = \text{loss}^{\text{prim}} + \sum_{j=1}^m \alpha_j \text{loss}^{\text{aux}_j}
\end{align}
Optionally, we can use activity regularization on $g_t$, such that it is either close to 1 or close to 0.  The regularization loss is:
\begin{align}
\text{loss}^{\text{reg}} = \lambda \sum_{t=1}^{n} \|\min(g_t, 1-g_t)\|_1
\end{align}

which discourages $g$ from ``sitting on the fence'' near~0.5.  Regularization loss is added to $\text{loss}^{\text{all}}$.


\section{Concrete instantiations of \system}
\label{sec:Concrete}

In this section, we present three concrete instantiations of \system, and their detailed architecture.

\subsection{Sentiment classification of code-switched passages}

In code-switched text, words from two languages are used.  Code switching is common in informal conversations and in social media where participants are multilingual users.  Usually, `imported' tokens are transliterated into a character set commonly used in the `host' language, making span language identification (the `language identification task') non-trivial.   Let the languages be $A$ and $B$ with vocabularies $D_A$ and $D_B$, respectively.  We are given a word sequence $(w_1, w_2, \ldots, w_N)$, where each word $w_i \in D_A \cup D_B$.  Our goal is to infer a sentiment label from $\{-1,0,+1\}$ for the whole sequence.  As stipulated in Section~\ref{sec:Arch}, labeled instances and classification techniques are readily available for text in single languages (auxiliary task) \citep{TangQFL2015TDLSTM,WangLZP2017TDParse}, but rare for code-switched text (primary task).  Recent sequence models for sentiment classification in one language essentially recognize the sentiments in words or short spans.  For such monolingual models to work well for code-switched text, the words and spans must be labeled with their languages, which is difficult and error-prone.
Combining signals from auxiliary models is not trivial because some words in language $A$ (e.g., `not' in English) could modify sentiments expressed by other words in language~$B$.

\subsubsection{Datasets:}
For the primary task we use the sentiment classification dataset of English-Spanish code-switched sentences \citep{vilares2015sentiment}. 
Each sentence has a human labeled sentiment class in $\{-1, 0 , 1\}$. The training and test sets contain 2,449 and 613 instances, respectively. 
We use two disjoint auxiliary task datasets.
For sentiment classification of English sentences, we use the Twitter dataset provided by Sentistrength\footnote{\url{http://sentistrength.wlv.ac.uk}}, which has 7,217 labeled instances.
For sentiment classification of Spanish sentences, we use the Twitter dataset by \cite{julio}, containing 4,241 labeled instances.

\subsubsection{Model description:} We use two LTSMs with 64 hidden units for the auxiliary task of English and Spanish sentiment classification.
The output of only the last step is fed into a fully connected layer with 3 units and softmax activation.
We use three separate fully connected layers for English, Spanish and English-Spanish tasks.
For the primary RNN which produces the gating signal, we use a bi-directional LSTM with 32 units. The gating signal is produced by adding a fully connected layer of 3 units with softmax activation on each step of the primary RNN. 
The word embeddings are initialized randomly and trained along with the model.

\subsection{POS tagging of code-switched sentences}

Our second application is part-of-speech (POS) tagging of code-switched sentences.  Unlike sentiment classification, here a label is associated with each token in the input sequence.  As in sentiment classification, we can use auxiliary models built from learning to tag monolingual sentences.  A word in language $A$ may have multiple POS tags depending on context, but context information may be provided by neighboring words in language~$B$.  E.g., in the sentence ``\textit{mein iss car ko like karta hu}" (meaning ``I like this car"), the POS tag of `like' depends on the Hindi text around it. The input sentence is again denoted by $(w_1, w_2, \ldots, w_N)$, with each word $w_i \in D_A \cup D_B$.  The goal is to infer a label sequence $(y_1, y_2, \ldots, y_N)$, where $y_i$ comes from the set of all POS tags over the languages.
To predict the POS tag at each word, we apply the fully connected layer to the composite hidden state.

\subsubsection{Datasets:}
For the primary task we use a Hindi-English code-switch dataset provided in a shard task of ICON'16
\citep{patra2018sentiment}. It contains sentences from Facebook, Twitter and Whatsapp. It has 19 POS tags, and the training and test sets have 2102 and 528 instances, respectively. 
For the auxiliary dataset of Hindi POS tagging, we use the data released by \cite{sachdeva2014hindi}, containing 14,084 instances with 25 POS tags.
For the auxiliary dataset of English POS tagging, we use the data released in a shared task of CoNLL 2000 \citep{tjong2000introduction} , containing 8,936 instances with 45 POS tags.

\subsubsection{Model description:}
We use two LTSMs with 64 hidden units for the auxiliary task of English and Hindi POS tagging.
For the primary task $h_t^{\text{comp}}$, each step is fed into a fully connected layer with 19 units with softmax activation.
For the two auxiliary tasks, $h_t^{\text{auxaux}_1}$ and $h_t^{\text{auxaux}_2}$ are fed into the two separate fully connected layers of 25 and 45 units at each word for Hindi and English tags, respectively.
For the primary RNN which produces the gating signal, we use a bi-directional LSTM with 32 units. The gating signal is produced by adding a fully connected layer of 3 units with softmax activation in each step of the primary RNN. The first two elements are the English and Hindi gates. The word embeddings are initialized randomly and trained along with the model.


\subsection{Target-dependent sentiment classification}

Our third application is target-dependent sentiment classification (TDSC), where we are given a sentence or short (assumed monolingual here) passage $(w_1, w_2, \ldots, w_N)$ with a designated span $[i,j]$ that mentions a named-entity target.  The primary task is to infer the polarity of sentiment expressed toward the target.
Our auxiliary task is the whole-passage sentiment classification, for which collecting labeled instances is easier \citep{go2009twitter,Sanders2011TwitterSentiment}.  The passage-level task has a label associated with the whole passage, rather than a specific target position.  E.g., in the tweet ``{\em I absolutely love listening to electronic music, however artists like Avici \& Tiesta copy it from others}'', the overall sentiment is positive, but the sentiments associated with both targets `Avici' and `Tiesta' are negative.

\subsubsection{Datasets:}

For the primary task i.e., target-dependent sentiment classification, we use dataset of the SemEval 2014 task 4 \citep{pontiki-EtAl:2014:SemEval}. It contains reviews from two domains -- Laptops (2,328 training and 638 testing instances) and Restaurants (3,608 training and 1120 testing instances). 
The dataset for its corresponding auxiliary task is Yelp2014, consisting of similar types of 183,019 reviews. All of these datasets have three classes $\{+1,0,-1\}$.

\subsubsection{Model description:} This application is another useful instantiation of \system, because (i) by training on whole-passage instances, the auxiliary RNN learns to identify the polarity of words and their local combinations (``not very good''), (ii) the primary RNN learns to focus on the span related to the target entity.

In order to infer the sentiment of the target entity, the model would have to find the regions which are related to the given target entity and use the information learned from the auxiliary task.  For qualitative analysis of our model, we will visualize the values of the scalar gates and find that words related to the target entity are passed through the auxiliary RNN and others are skipped/blocked.
Here we implement \shortname{} on top of TD-LSTM  \citep{tang2015effective}.  We can think of it as two separate instances, for the left and right context spans of the target.

For the primary task i.e., target-dependent sentiment classification, we use two separate RNNs to produce the gating signals and capture primary features which are not captured by the auxiliary RNN. Since there is only one auxiliary task, the controller either skips the RNN at a particular word or it may pass it to the auxiliary RNN. 
Given an input sentence and a target entity, we split the sentence at the position of the target entity. We input the left half of the sentence to the left primary and auxiliary RNN and the right half to the right auxiliary and primary RNN. Rather than pooling the states, we do a weighted pool where we sum the hidden states after multiplying with the gate values.  \textcolor{black}{Similar to TD-LSTM, The pooling of left and right is done separately and then concatenated.}

For the auxiliary task i.e., sentiment classification of the complete sentence, \textcolor{black}{we use a left RNN and a right RNN so that we can couple them with the two primary RNNs} . We run both the left and right auxiliary  RNNs on the input text and the reversed input text respectively. At each token of the input sentence we concatenate the hidden state of the left RNN and the right RNN. For the sentiment classification of the complete sentence, we take an average pool the features of all the token positions and pass it to a fully connected layer with softmax activation for classification of the sentiment score.


\section{Comparative Evaluation}
\label{sec:Expt}

In this section, we first describe baseline approaches, followed by a comparison between them and \shortname{} over the three concrete tasks we set up in Section~\ref{sec:Concrete}.

\subsection{Baseline Methods}

Here we briefly describes several MTL-based baselines (along with their variants) used in all three applications mentioned in Section~\ref{sec:Concrete}.   A few other application-specific baselines are mentioned in Section~\ref{sec:result}.

\subsubsection{LSTM (no MTL):} LSTM trained on just the primary task.

\subsubsection{Hard Parameter Sharing (HardShare):}
This model \citep{Caruana93} uses the same LSTM for all the tasks along with the separate task-specific classifiers. We show results for both 1-layer LSTM \textbf{(HardShare~1L)} and 2-layer LSTM \textbf{(HardShare~2L)}.

\subsubsection{Low Supervision (LowSup):} 
This model \citep{Sogaard2016MTLNLP} uses a 2-layer LSTM, where the auxiliary tasks are trained on the lower layer LSTM, and the primary task is trained on the higher layer LSTM. Here we show results on two schemes: 1)~\textbf{LowSup Share:}  same LSTM is used for all auxiliary tasks, 2)~\textbf{LowSupConcat:} Outputs of separate layer-1 LSTM concatenated. 

\subsubsection{Cross-stitch Network (XStitch):}
This model \citep{Misra2016CrossStitch} has separate LSTM for each task. The amount of information shared to the next layer is controlled by trainable scalar parameters. We show results for both 1-layer LSTM \textbf{(XStitch~1L)} and 2-layer LSTM \textbf{(XStitch~2L)}.

\subsubsection{Sluice Network (Sluice):} This is an extension \citep{ruder2017learning} of cross-stitch network, where the outputs of the intermediate layers are fed to the final classifier.  Here, rather than applying the stitch module on the LSTM outputs at a layer, they split the channels of each LSTM and apply the stitch module.

\begin{table}[!t]
\centering\adjustbox{max width=\hsize}{
\begin{tabular}{|l|c|c|c|c|}
\hline
\multicolumn{1}{|c|}{\textbf{Model}}                          & \textbf{Accuracy} & \textbf{Macro F\textsubscript{1}}    & \textbf{Precision} & \textbf{Recall} \\ \hline
LSTM (No MTL)                 & 59.22    & 58.34 & 58.86     & 58.01        \\  
HardShare 1L  & 60.36    & 59.65 & 60.28     & 59.23       \\  
HardsShare 2L & 57.10     & 55.21 & 56.59     & 54.88       \\
LowSupShare & 55.95 & 54.42 & 55.42 & 54.23 \\
LowSupConcat & 56.61 & 55.84 & 57.22&55.25  \\
XStitch 1L   & 59.71    & 58.59 & 59.94     & 58.08        \\  
XStitch 2L  & 56.44    & 48.31 & 58.76     & 51.35       \\  
Sluice         & 58.56    & 57.71 & 57.61     & 58.18        \\ 
PSP-MTL & 59.22 & 59.23 &  59.04 & 59.62 \\
SSP-MTL & 59.71& 59.75& 59.45& 60.33 \\
Meta-MTL & 57.59 & 54.56 &58.39 & 54.24\\
Coupled RNN & 56.77 & 55.69 & 55.85 &55.70 \\\hline
\rowcolor{green!15}
\system~1L & \textbf{63.30}     & \textbf{62.36} & \textbf{63.35}     & \textbf{61.83}  \\ 
\rowcolor{green!5}
\system~2L & 61.99    & 61.23 & 61.79     & 61.13       \\ \hline 
\end{tabular}}
\caption{Accuracy of the competing models for the sentiment classification of code-switched passages. F\textsubscript{1} score, Precision and Recall are macro averaged. }  \label{tab:sentiment}
\end{table}

\subsubsection{Shared-private sharing scheme:} In this architecture there is a common shared LSTM over all tasks and separate LTSMs for each task. We show results on two schemes 1)~parallel shared-private sharing scheme \textbf{(PSP-MTL)} as described by \cite{liu2017adversarial}, where the outputs of the private LSTM and shared LSTM are concatenated; and 2)~stacked shared-private sharing scheme \textbf{(SSP-MTL)} as described by \cite{Chen2018MetaMTL}, where output of the shared LSTM is concatenated with the sentence which is fed to the private LSTM.

\subsubsection{Meta multi-task learning (Meta-MTL):} 
In this model \citep{Chen2018MetaMTL}, the weight of the task specific LSTM is a function of a vector produced by the Meta LSTM which is shared across all the tasks. 

\subsubsection{Coupled RNN:} This model \citep{liu2016recurrent} has an LSTM for each task which uses the information of the other task.


\begin{table}[!t]
\centering\adjustbox{max width=\hsize}{
\begin{tabular}{|l|c|c|c|c|} \hline
\multicolumn{1}{|c|}{\textbf{Model}} & \textbf{Accuracy} & \textbf{Macro F\textsubscript{1}}    & \textbf{Precision} & \textbf{Recall} \\ \hline
LSTM (no MTL) & 61.46 & 40.55 & 47.18    & 40.84 \\
HardShare 1L  & 62.19 & 40.55 & 45.00    & 39.18 \\
HardShare 2L& 62.10  & 42.33 & 47.77 & 40.87 \\
LowSupShare & 61.99 & 43.62 & 46.88  & 42.53 \\
LowSupConcat & 62.66 & 43.82 & 50.17 &41.75 \\
XStitch 1L   & 63.28 & 44.39 & 
\cellcolor{green!15} \textbf{54.51} & 42.11 \\
XStitch 2L  & 62.88 & 41.78 & 48.03 & 39.75 \\
Sluice  & 60.90  & 40.81 & 43.95 & 40.35 \\
PSP-MTL & 62.93 & 41.88 & 47.85 & 39.98 \\
SSP-MTL & 62.90 & 37.63 & 46.36 & 35.49 \\
Meta-MTL & 62.25 & 41.33 & 47.4 & 39.77 \\
Coupled RNN & 62.44 & 41.91 & 52.73 & 40.20 \\\hline
\rowcolor{green!15}
\system~1L  & \textbf{64.29} & \textbf{47.58} & 
\cellcolor{green!5} 53.51 & \textbf{45.48} \\
\system~2L  & 63.13 & 45.75 & 51.34 & 43.65 \\\hline
\end{tabular}}
\caption{Accuracy of the competing models for the POS tagging of code-switched sentences.  F\textsubscript{1} score, Precision and Recall are macro averaged. }   \label{tab:postag} 
\end{table}

\subsection{Experimental Results}\label{sec:result}

Tables \ref{tab:sentiment}, \ref{tab:postag} and \ref{tab:tdsc} show the results for \system\ and other MTL baselines for three applications mentioned in Section \ref{sec:Concrete}. We observe a significant improvement of \system\ over all other models. For a fair comparison with multi-layer LSTM MTL models, we show results of \system\ with two LSTM layers. This is because some models like Sluice and LowSup can only be implemented with $>1$ LSTM layers. We see that for three applications, all the 2-layer LSTM based models have worse performance compared to the single layer LSTM models. However, \system\ with two layers outperforms all other 2-layer LSTM models. 
The Coupled RNN and the Meta-MTL are defeated by \system\ because due to scalar gates, \system\  has fewer degrees of freedom for manipulation of information, which helps in learning with less data. \system\ beats XStitch and Sluice as they do not have any sharing of information at a granularity of words.

We also compare \system\ with some single-task baselines in Table \ref{tab:tdsc}. These methods are designed particularly for target-specific sentiment classification.
TD-LSTM \citep{tang2015effective} is the no-MTL baseline for the task of TDSC as in this case, \system\ is implemented on top of TDSC. In the TD-LSTM + Attention model \citep{wang2016attention}, attention score at each token is computed which is used to do a weighted pooling of the hidden states. In Memory network (MemNet) \citep{tang2016aspect}, multiple modules of memory are stacked and the initial key is the target entity. 


\begin{table}
\centering\adjustbox{max width=\hsize}{
\begin{tabular}{|l|c|c|c|c|}
\hline
\multirow{2}{*}{\textbf{Model}} & \multicolumn{2}{c|}{\textbf{Laptop}}  & \multicolumn{2}{c|}{\textbf{Restaurant}} \\ \cline{2-5} 
& \textbf{Accuracy} & \textbf{F\textsubscript{1}} & \textbf{Accuracy}       & \textbf{F\textsubscript{1}}        \\ \hline
TD-LSTM & 71.38 & 68.42 & 78.00 & 66.73 \\
TD-LSTM + Att. & 72.14 & 67.45 & 78.89 & 69.01 \\
MemNet & 70.33 & 64.09 & 78.16  & 65.83 \\
\hline
HardShare 1L  & 72.27 & 66.71 & 78.66    & 66.08 \\
HardShare 2L & 70.72 & 63.70 & 78.13 & 66.16 \\
LowSupShare &71.65 &65.74 & 79.64 & 68.46 \\
XStitch 1L  & 71.81 & 65.5 & 79.02 & 68.55 \\
XStitch 2L & 73.05 & 67.63 & 78.93 & 68.24 \\
Sluice &  71.50 & 66.10 & 78.84 & 69.62 \\
PSP-MTL & 71.65 & 65.45 & 79.55 &  68.75\\
SSP-MTL & 70.87 & 65.93 & 79.11 & 69.32 \\
Meta-MTL & 71.34 & 66.19 & 78.66 & 68.17 \\
Coupled RNN & 71.34 & 64.68  & 79.19 & 65.98 \\\hline
\rowcolor{green!10}
GIRNet 1L                     & {74.92} & {69.67} & \textbf{82.41} & \textbf{74.35}     \\
\rowcolor{green!10}
GIRNet 2L & \textbf{75.86} & \textbf{71.39} & 80.18 & 69.14 \\\hline
\end{tabular}}
\caption{Accuracy and macro F1 of the competing models for the target-dependent sentiment classification.} \label{tab:tdsc}  
\end{table}

\subsection{Visualization }

To get insight into \shortname's success, we studied the scalar gate values at each word of input sentences.
Table \ref{tab:tdscheatmap} shows scalar gate values for a few TDSC instances.  We see that words associated with the target entity get larger gate values, which is particularly beneficial for multiple entities and diverse sentiments.  Table \ref{tab:sentimentheatmap} shows gate values for sentiment classification task on code-switched passages.  Here we do not visualize the absolute gate values because state vectors of the two RNNs can have widely diverse magnitudes.  Instead, we visualize the change in the value of the gates on switching the sentiment-charged words from one language to the other.  Green and blue colors represent the magnitude of drop in the value of gate corresponding to the auxiliary RNN for Spanish and English tasks, respectively.

\begin{table}[!t]
\centering
\begin{tabular}{|l|p{60mm}|} \hline
Target & Gate heatmap \\ \hline
Soup         &    \yhl{1,0.992157,0.992157}{the} \yhl{1,0.980392,0.980392}{service} \yhl{1,0.996078,0.996078}{is} \yhl{1,0.552941,0.552941}{great}, \yhl{1,0.325490,0.325490}{my} \yhl{1,1.000000,1.000000}{\underline{soup}} \yhl{1,0.309804,0.309804}{always} \yhl{1,0.337255,0.337255}{arrives} \yhl{1,0.282353,0.282353}{nice} \yhl{1,0.894118,0.894118}{and} \yhl{1,0.623529,0.623529}{hot}. \\ \hline
Appetizers         & \yhl{1,1.000000,1.000000}{\underline{appetizers}} \yhl{1,0.584314,0.584314}{are} \yhl{1,0.278431,0.278431}{ok}, \yhl{1,0.984314,0.984314}{but} \yhl{1,0.996078,0.996078}{the} \yhl{1,1.000000,1.000000}{service} \yhl{1,1.000000,1.000000}{is} \yhl{1,0.996078,0.996078}{slow}. \\ \hline
Service        & \yhl{1,1.000000,1.000000}{appetizers} \yhl{1,0.996078,0.996078}{are} \yhl{1,0.870588,0.870588}{ok}, \yhl{1,0.352941,0.352941}{but} \yhl{1,0.290196,0.290196}{the} \yhl{1,1.000000,1.000000}{\underline{service}} \yhl{1,0.537255,0.537255}{is} \yhl{1,0.317647,0.317647}{slow}.
\\ \hline
\end{tabular}
\caption{Gating heatmaps of Target-dependent sentiment classification.  In each case, words associated with the target get the largest gate values.}\label{tab:tdscheatmap}
\end{table}

\begin{table}[!t]
\centering
\begin{tabular}{|p{80mm}|} \hline
\yhl{1.000000,1.000000, 1}{The} \yhl{1.000000,1.000000, 1}{quality} \yhl{1.000000,1.000000, 1}{of} \yhl{1.000000,1.000000, 1}{the} \yhl{1.000000,1.000000, 1}{laptop} \yhl{1.000000,1.000000, 1}{is} \textcolor{white}{\yhl{0.076792,0.076792, 1}{good}} \yhl{1.000000,1.000000, 1}{and} \yhl{1.000000,1.000000, 1}{el} \yhl{1.000000,1.000000, 1}{servicio} \yhl{1.000000,1.000000, 1}{de} \yhl{1.000000,1.000000, 1}{la} \yhl{1.000000,1.000000, 1}{compa\~n\'ia} \yhl{1.000000,1.000000, 1}{es} \yhl{0.270477,1,0.270477}{genial} \\ \hline
\yhl{1.000000,1.000000, 1}{i} \yhl{0.615114,0.615114, 1}{love} \yhl{1.000000,1.000000, 1}{this} \yhl{1.000000,1.000000, 1}{city} \yhl{1.000000,1.000000, 1}{because} \yhl{1.000000,1.000000, 1}{el} \yhl{1.000000,1.000000, 1}{clima} \yhl{1.000000,1.000000, 1}{es} \yhl{0.329823,1,0.329823}{bueno}  \yhl{1.000000,1.000000, 1}{y} \colorbox[rgb]{1.000000,1.000000, 1}{la} \colorbox[rgb]{1.000000,1.000000, 1}{gente} \colorbox[rgb]{1.000000,1.000000, 1}{es} \colorbox[rgb]{0.667556,1,0.667556}{amable} \\ \hline
\end{tabular}
\caption{Gating heatmap for the sentiment classification of code-switched passages. Green and blue colors correspond to the gates responsible for the Spanish auxiliary RNN and English auxiliary RNN, respectively. 
}
\label{tab:sentimentheatmap}
\end{table}

\section{Conclusion}
\label{sec:End}

Sequence labeling tasks are often applied to multi-domain (such as code-switched) sequences. But labeled multi-domain sequences are more difficult to collect compared to single-domain (such as monolingual) sequences.  We therefore need \emph{sequence MTL}, which can train auxiliary sequence models on single-domain instances, and learn, in an unsupervised manner, how to interleave composite sequences by drawing on the best auxiliary sequence model cell at each token position.  Our proposed \shortname\ precisely serves this need.  We tested our model on three concrete applications and obtained larger accuracy gains compared to other MTL architectures.  Inspection of the interleaving positions proves that \system\ learns to identify the intended token-level signals for effective state composition.

\medskip\paragraph*{Acknowledgement:} Partly supported by Microsoft Research India travel grant, IBM, Early Career Research Award (SERB, India), and the Center for AI, IIIT Delhi, India.

\clearpage
\bibliographystyle{aaai_camera}
\begin{small}
\bibliography{voila,mttdsc,newbib}
\end{small}

\end{document}